\documentclass[10pt,twocolumn,letterpaper]{article}

\usepackage{cvpr}
\usepackage{times}
\usepackage{epsfig}
\usepackage{graphicx}
\usepackage{amsmath}
\usepackage{amssymb}

\usepackage{amsmath}
\usepackage{multirow}
\usepackage{url}
\usepackage{array}

\usepackage{wrapfig}
\usepackage{subfig}
\usepackage{float}
\usepackage{booktabs}

\usepackage[breaklinks=true,bookmarks=false]{hyperref}

\cvprfinalcopy 

\def\cvprPaperID{****} 

\begin{document}

\title{IVUS-Net: An Intravascular Ultrasound Segmentation Network}

\author{Ji Yang, Lin Tong, Mehdi Faraji\thanks{Corresponding author.}, Anup Basu\\
Department of Computing Science,\\ University of Alberta, Canada\\
{\tt\small \{jyang7, ltong2, faraji, basu\}@ualberta.ca}}

\maketitle

\begin{abstract}
\textbf{I}ntra\textbf{V}ascular \textbf{U}ltra\textbf{S}ound (IVUS) is one of the most effective imaging modalities that provides assistance to experts in order to diagnose and treat cardiovascular diseases.
We address a central problem in IVUS image analysis with Fully Convolutional Network (FCN): automatically delineate the lumen and media-adventitia borders in IVUS images, which is crucial to shorten the diagnosis process or benefits a faster and more accurate 3D reconstruction of the artery. Particularly, we propose an FCN architecture, called IVUS-Net, followed by a post-processing contour extraction step, in order to automatically segments the interior (lumen) and exterior (media-adventitia) regions of the human arteries. We evaluated our IVUS-Net on the test set of a standard publicly available dataset containing 326 IVUS B-mode images with two measurements, namely Jaccard Measure (JM) and Hausdorff Distances (HD). The evaluation result shows that IVUS-Net outperforms the state-of-the-art lumen and media segmentation methods by 4\% to 20\% in terms of HD distance. IVUS-Net performs well on images in the test set that contain a significant amount of major artifacts such as bifurcations, shadows, and side branches that are not common in the training set. Furthermore, using a modern GPU, IVUS-Net segments each IVUS frame only in 0.15 seconds. The proposed work, to the best of our knowledge, is the first deep learning based method for segmentation of both the lumen and the media vessel walls in 20 MHz IVUS B-mode images that achieves the best results without any manual intervention. Code is available at \url{https://github.com/Kulbear/ivus-segmentation-icsm2018}.
\end{abstract}

\section{Introduction}
\label{sec:intro}

Convolutional Neural Networks (CNNs) play an important role in visual image recognition. In the past few years, CNNs have achieved promising results in image classification \cite{krizhevsky2012imagenet,simonyan2014very,srivastava2015highway,he2016deep,huang2017densely} and semantic segmentation \cite{ciresan2012deep,long2015fully,chen2016deeplab,rajpurkar2017chexnet,peng2017large}. Fully Convolutional Networks (FCNs) \cite{long2015fully} have become popular and used to solve the problem of making dense predictions at a pixel level. There are two major differences between FCN and the type of CNNs which are primarily designed for classification \cite{krizhevsky2012imagenet,simonyan2014very,szegedy2015going}. First, FCN does not have fully-connected layers therefore can accept any arbitrary size of inputs. Secondly, FCN consists of an encoder network that produces embedded feature maps that are followed by a decoder network to expand and refine the feature maps outputted by the encoder. Skip connections are also common in the architecture to connect corresponding blocks in the encoder and decoder \cite{badrinarayanan2017segnet,drozdzal2016importance,chen2016deeplab}.

Segmentation of the acquired IVUS images is a challenging task since IVUS images usually comes with artifacts. Particularly, a successful separation of the interior (lumen) and exterior (media) vessel walls in IVUS images plays a critical role to diagnose cardiovascular diseases. It also helps building the 3D reconstruction of the artery where the information of the catheter movements has been provided using another imaging modality such as X-Ray. The segmentation of IVUS images has been a well-investigated problem from a conventional perspective where numerous ideas and approaches of computer vision and image processing such as in \cite{mendizabal2013segmentation,mendizabal2016physics,taki2008automatic,zhu2011snake,unal2008shape} have been employed. One of the best segmentation results have been achieved in a very recent work \cite{faraji2018segmentation} where authors proposed a two-fold IVUS segmentation pipeline based on traditional computer vision methods \cite{faraji2015extremal,faraji2015erel}. Although no learning method was used, it outperforms existing methods from both the accuracy and efficiency perspective. Although the reported performance of \cite{faraji2018segmentation} is very close to the ground truth label (0.30mm error of the segmented lumen and 0.22mm error of the segmented media from the gold standard), we believe that the deep learning technique has the potential to perform better.
\begin{figure*}[t]
	\includegraphics[width=1\textwidth]{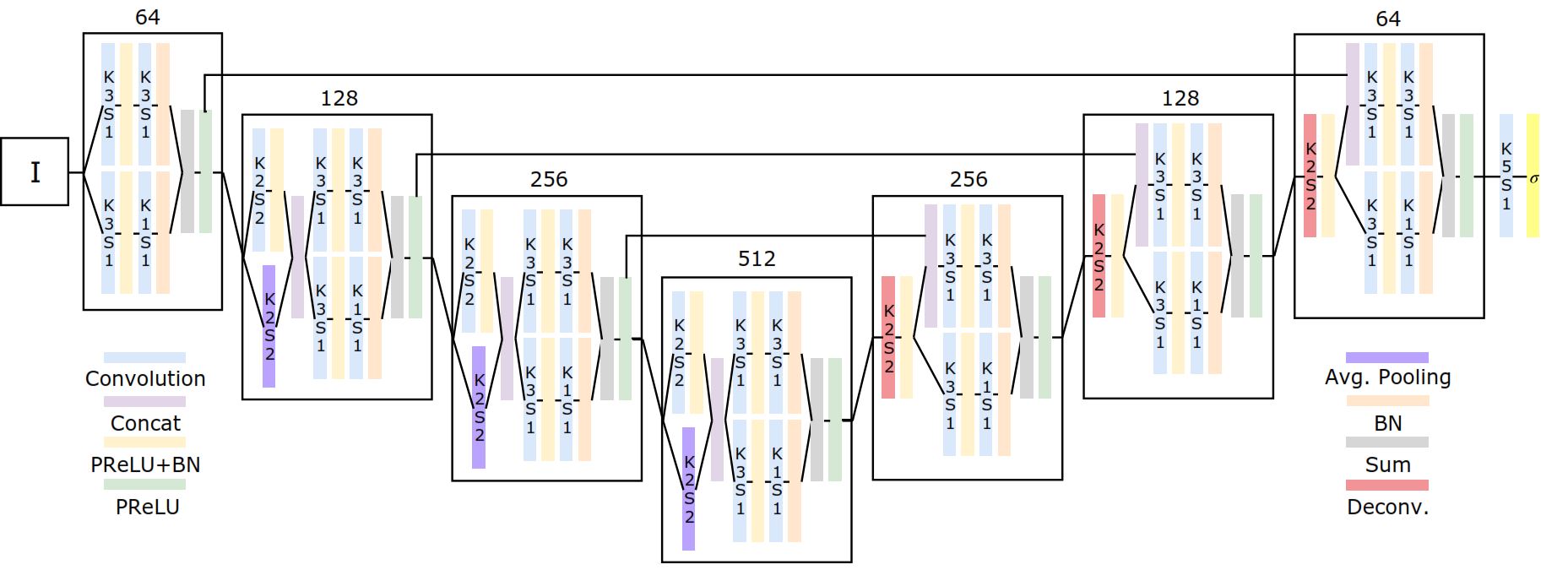}
	\caption{The IVUS-Net architecture. Every convolutional layer in the same block has the same output depth as labeled on the top of the block.
		The figure should be read from left to right as we omitted arrow heads to save the space.}
	\label{fig:arch}
\end{figure*}
In this paper, we propose an FCN-based pipeline that automatically delineates the boundary of the lumen and the media vessel walls. The pipeline contains two major components: a carefully designed FCN for predicting a pixel-wise mask which is called IVUS-Net, followed by a contour extraction post-processing step. In addition, the FCN is trained from scratch without relying on any pre-trained weights. We evaluated the proposed IVUS-Net on the test set of a publicly available IVUS B-mode benchmark dataset \cite{balocco2014standardized} which contains 326 20MHz IVUS images consist of various artifacts such as motion of the catheter after a heart contraction, guide wire effects, bifurcation and side-branches. Two standard metrics, namely, Jaccard Measure (JM), alternatively called Intersection over Union (IoU), and Hausdorff Distance (HD) were used for evaluation. 

\label{subsec:contribution}
The contributions of the proposed work can be summarized as follows:
\begin{itemize}
	\item We propose a pipeline based on a FCN followed by a post-processing contour extraction to automatically delineate the lumen and media vessel walls.
	\item We show that the proposed work outperforms the current state-of-the-art studies over a publicly available IVUS benchmark dataset \cite{balocco2014standardized} which contains IVUS images with a significant amount of artifacts. This shows that the proposed work has the potential to be generalized to other IVUS benchmarks as well.
	\item To the best of our knowledge, there is no previous work based on deep architecture that can produce segmentation for both the lumen and media vessel walls in B-mode IVUS images.
\end{itemize}

The rest of the paper is organized as the follows: Section \ref{sec:method} contains a detailed description of our proposed work. In Section \ref{sec:experiments}, we demonstrate multiple experiments that reinforce our contribution. Finally, we conclude the work briefly in Section \ref{sec:conclusion}.

\section{Proposed Method}
\label{sec:method}

In this section, we first introduce the dataset we used to train the deep model. Then, we present the architecture, IVUS-Net, that produces binary prediction mask for either the lumen or media area, followed by a contour extraction step to delineate the vessel wall.

\subsection{Dataset}
\label{subsec:dataset}
We used a publicly available IVUS dataset \cite{balocco2014standardized} that contains two sets (train and test) of IVUS gated frames using a full pullback at the end-diastolic cardiac phase from 10 patients. Each frame has been manually annotated by four clinical experts. The train and test sets consist of 109 and 326 IVUS frames, respectively. Also, test set contains a large number of IVUS artifacts including bifurcation (44 frames), side vessel (93 frames), and shadow (96 frame) artifacts. The remaining 143 frames do not contain any artifacts except for plaque.
\begin{figure*}[t]
	\centering
	\begin{tabular}{cc}
		\includegraphics[width=0.45\textwidth]{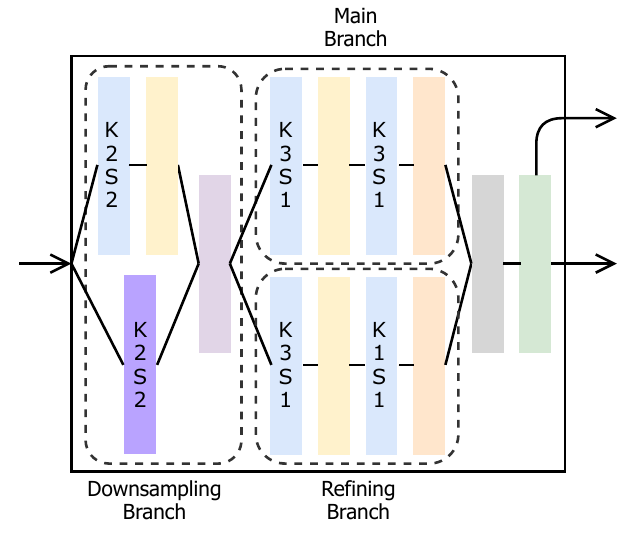} & \includegraphics[width=0.45\textwidth]{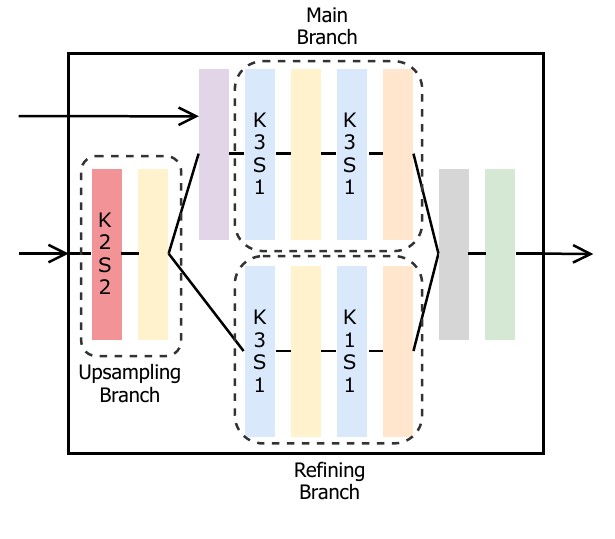}\\
		(a) & (b)\\
	\end{tabular}
	\caption{A detailed illustration of the encoding block and the decoding block. Note the first encoding block does not have the downsampling branch, therefore the main branch and refining branch will directly accept the raw image as the input. (a) An encoding block with downsampling branch, followed by the main branch and the refining branch. (b) A typical decoding block that accept feature map from both the previous block and the skip-connection.}
	\label{fig:blocks}
\end{figure*}
\subsection{IVUS-Net}
\label{subsec:ivusnet}

IVUS-Net is designed based on fully convolutional network (FCN) \cite{long2015fully}, with inspirations from aggregated, multi-branch architectures such as ResNeXT \cite{xie2017aggregated} and the Inception model \cite{szegedy2015going}. Both SegNet \cite{badrinarayanan2017segnet} and U-Net \cite{ronneberger2015u} can be considered as a base version of our proposed work according to the network architecture design. It has two major components:
\begin{enumerate}
	\item An encoder network that can downsample and process the input to produce a low-resolution deep feature map.
	\item A decoder network that can restore the resolution of the deep feature map outputted by the encoder network towards original size.
\end{enumerate}  
The output feature map is sent to one more convolutional layer followed by a sigmoid activation to produce the final result.

The encoder network contains 4 encoding blocks where the decoder network contains 3 decoding blocks. Each decoding block receives feature map from its previous block and extra information from the encoder network by skip-connections. The entire architecture is therefore symmetric as shown in Fig.~\ref{fig:arch}. There are minor differences among the blocks in the architecture. We give a brief illustration for the design and also show the intuitions behind.

Except for the first encoding block, each encoding block contains a downsampling branch that downsamples the input feature map, then followed by a two-branch convolution path, as shown in Fig.~\ref{fig:blocks}(a). We build and expand downsampling branches in order to avoid losing information due to using the pooling. In fact, the downsampling branch facilitates reducing the spatial resolution of the input. It employs a 2-by-2 average pooling layer and a 2-by-2 convolutional layer with a stride of 2 at the same time, and finally concatenate the two outputs together, this aggregation idea is similar to \cite{szegedy2015going,xie2017aggregated}.

After the downsampled, aggregated feature map outputs by the downsampling branch is passed to two subsequent branches, namely the refining branch and the main branch.
First, we follow the design in \cite{badrinarayanan2017segnet,ronneberger2015u} to include a branch with consecutive convolutional layers followed by activation and batch normalization, here we call it ``main branch''. 
A recent trend is to use small kernel size for the feature map refinement \cite{chen2016deeplab,peng2017large}. So we intentionally design a ``refining branch'' that has one convolutional layer with a 3-by-3 kernel size followed by a convolutional layer with a 1-by-1 kernel size produces similar but refined feature map. 
The outputs from the main branch and refining will be summed up and pass to the next block and its corresponding decoding block.

Decoding blocks need a slightly different configuration, as shown in Fig.~\ref{fig:blocks}(b). Every decoding block receives the feature map from both its previous block and its corresponding encoding block. Only the feature map received from the previous block is upsampled by a 2-by-2 deconvolution and then concatenated with the feature map from its corresponding encoding block.
Note that this concatenated feature map will only be passed to the main branch, where the refining branch handles the upsampled feature map only.

The activation used in the IVUS-Net is the \textbf{P}arametric \textbf{Re}ctified \textbf{L}inear \textbf{U}nit (PReLU) \cite{he2015delving}.
\begin{equation}
\textmd{PReLU}(x) = \max(0, x) - \alpha \max(0, -x)
\end{equation}
Compared with the ordinary ReLU activation, PReLU allows a part of the gradients flow through when the neuron is not activated, where ReLU only passes gradients when the neuron is active. As suggested in \cite{he2015delving,xu2015empirical}, PReLU outperforms ReLU in many benchmarks and also has a more stable performance.

Finally, the output feature map from the last decoding block is refined by a 5-by-5 convolutional layer, which is experimentally proved to be helpful on improving performance. As we want IVUS-Net to produce binary masks, the last activation is a sigmoid function.

\subsection{Post-processing}
Generally, as it has been proposed in \cite{faraji2018segmentation}, since the shape of the lumen and media regions of the vessel are very similar to conic sections, representing the predicted masks by fitting an ellipse on the masks can increase the accuracy of the segmentations. Therefore, we follow the same process explained in \cite{faraji2018segmentation} to post-process the predicted masks in order to extract the final contours.

\section{Experiments}
\label{sec:experiments}

The evaluation is based on a publicly available IVUS B-mode dataset \cite{balocco2014standardized}, which has been widely used in the IVUS segmentation literature \cite{faraji2018segmentation,downe2008segmentation,mendizabal2013segmentation,zhu2011snake,unal2008shape}. There are 109 images in the training set, 326 images in the test set and no official validation set is provided.
Models are trained end-to-end, based on only the given dataset without involving any other external resources such as extra training images and pre-trained model weights. 
Two metrics are used for the evaluation, namely Jaccard Measure (JM) and Hausdorff Distance (HD). 
The Jaccard Measure, sometimes called Intersection over Union, 
is calculated based on the comparison of the automatic segmentation from the pipeline ($R_{{pred}}$) and the manual segmentation delineated by experts ($R_{{true}}$).
\begin{equation}
\textmd{JM} = \frac{R_{{pred}} \cap R_{{true}}}{R_{{pred}} \cup R_{{true}}}
\end{equation}
The Hausdorff Distance between the automatic ($C_{{pred}}$) and manual ($C_{{true}}$) curves is the ``greatest distance of all points belonging to $C_{{pred}}$ to the closest point'' \cite{faraji2018segmentation} in $C_{{true}}$ and is defined as follows: 
\begin{equation}
\textmd{HD} = \max \left\{ d(C_{{pred}}, C_{{true}}), d(C_{{true}}, C_{{pred}}) \right\} 
\end{equation}
\begin{table}[t]
	\centering
	\renewcommand{\arraystretch}{1.2}
	\caption{Data Augmentation and Refining Branch Validation.}
	\subfloat[Data augmentation evaluation result]{
		\label{table:aug-compare}
		\begin{tabular}{@{}c|cc|cc@{}}
			\toprule
			& \multicolumn{2}{c}{Lumen} & \multicolumn{2}{c}{Media} \\
			& Jacc.          & Acc.        & Jacc.          & Acc.        \\ \midrule
			Aug. Data & 0.86        & 98.6        & 0.84        & 96.8        \\
			Orig. Data  & 0.83        & 97.6        & 0.79        & 96.1        \\ \bottomrule
		\end{tabular}
	}
	\quad
	\subfloat[Refining branch evaluation result]{
		\label{table:tri-compare}
		\begin{tabular}{@{}c|cc|cc@{}}
			\toprule
			& \multicolumn{2}{c}{Lumen} & \multicolumn{2}{c}{Media} \\
			& Jacc.          & Acc.        & Jacc.          & Acc.        \\ \midrule
			W/ Ref Path & 0.86        & 98.6        & 0.84        & 96.8        \\
			W/O Ref Path  & 0.84        & 97.9        & 0.80        & 96.0        \\ \bottomrule
		\end{tabular}
	}
\end{table}
\subsection{Data Augmentation}
\label{subsec:data-aug}

The training set contains only 109 images, which is considered as a relatively small training set for training a deep model from scratch. We then employ data augmentation on all the available training images. The augmentation is twofold. First, every original IVUS image and its corresponding ground truth masks are flipped (1) left to right, (2) up to down, and (3) left to right then up to down, to generate three new image-mask pairs. Secondly, we add heavy noises to input images. The methods we use to add noises to the input image include giving additive Gaussian noise, as suggested by \cite{zhang2016understanding}, or converting the input image to entirely black. No modification is done on the ground truth masks. The effectiveness of data augmentation is discussed in Section~\ref{subsec:eff-data-aug}.

\subsection{Training the Model}
\label{subsec:train}

All the models are trained and evaluated on a computer with a Core i7-8700K processor, 16GB of RAM, and a GTX 1080 8GB graphics card. 
Training a model from scratch generally takes less than 2 hours to complete. To make the training faster and use a relatively large batch size, we downsized every frame of the dataset by a factor of 0.5.

We implement IVUS-Net with TensorFlow \cite{tensorflow2015whitepaper}. 
The weights in the model are all initialized randomly.
Then we train the model with Adam optimizer \cite{kingma2014adam}. 
The learning rate is set to be 0.0001 with no decay scheme. 
The augmented training set is used to train each model for 96 epochs, with a batch size of 6 and 144 iterations in total for each epoch. 
Note that we need two groups of models to predict the lumen area and the media area since the output activation is a sigmoid function:
\begin{equation}
\sigma(x) = \frac{1}{1 + e^{-x}}
\end{equation}
For training each model, we randomly select 10 original IVUS images as the validation set to monitor the average Jaccard Measure without extracting contours.
The given prediction by a single model is a probability map that has equal dimensions to the input image size. We follow the ensemble practice in \cite{ciresan2012deep} to produce the final result.
\begin{table*}[t]
	\centering
	\caption{Performance of the proposed IVUS-Net with contour extraction. Measures represent the mean and standard deviation evaluated on 326 frames of the dataset\cite{balocco2014standardized} and categorized based on the presence of a specific artifact in each frame. The evaluation measures are Jaccard Measure (JM) and Hausdorff Distance (HD).}
	\label{table:experiment}
	\begin{tabular}{@{}cc|cc|cc@{}}
		\toprule
		&                 & \multicolumn{2}{c}{Lumen}                   & \multicolumn{2}{c}{Media}                   \\ 
		&                 & JM                   & HD                   & JM                   & HD                   \\
		\midrule
		All & Proposed  & \textbf{0.90 (0.06)} & \textbf{0.26 (0.25)} & \textbf{0.86 (0.11)} & \textbf{0.48 (0.44)} \\
		& Faraji et al. \cite{faraji2018segmentation}  & 0.87 (0.06)          & 0.30 (0.20)          & 0.77 (0.17)          & 0.67 (0.54)          \\
		& Downe et al. \cite{downe2008segmentation}    & 0.77 (0.09)          & 0.47 (0.22)          & 0.74 (0.17)          & 0.76 (0.48)          \\
		& Exarchos et al. \cite{balocco2014standardized} & 0.81 (0.09)          & 0.42 (0.22)          & 0.79 (0.11)          & 0.60 (0.28)          \\
		\midrule
		No Artifact     & Proposed & \textbf{0.91 (0.03)} & \textbf{0.21 (0.09)} & \textbf{0.92 (0.05)} & \textbf{0.27 (0.23)}          \\
		& Faraji et al. \cite{faraji2018segmentation}   & 0.88 (0.05)          & 0.29 (0.17)          & 0.89 (0.07)          & 0.31 (0.23) \\
		\midrule
		Bifurcation     & Proposed & \textbf{0.82 (0.11)} & 0.50 (0.58)         & \textbf{0.78 (0.11)}          & 0.82 (0.60)          \\
		& Faraji et al. \cite{faraji2018segmentation}   & 0.79 (0.10)          & 0.53 (0.34) & 0.57 (0.13)          & 1.22 (0.45)          \\
		& Downe et al. \cite{downe2008segmentation}    & 0.70 (0.11)          & 0.64 (0.27)          & 0.71 (0.19)          & 0.79 (0.53)          \\
		& Exarchos et al. \cite{balocco2014standardized} & 0.80 (0.09)          & \textbf{0.47 (0.23)}          & 0.78 (0.11) & \textbf{0.63 (0.25)} \\
		\midrule
		Side Vessels    & Proposed & \textbf{0.90 (0.04)} & \textbf{0.23 (0.12)} & \textbf{0.83 (0.14)} & \textbf{0.59 (0.49)}          \\
		& Faraji et al. \cite{faraji2018segmentation}   & 0.87 (0.05)          & 0.24 (0.11)          & 0.73 (0.60)          & 0.74 (0.18)          \\
		& Downe et al. \cite{downe2008segmentation}    & 0.77 (0.08)          & 0.46 (0.19)          & 0.74 (0.16)          & 0.76 (0.47)          \\
		& Exarchos et al. \cite{balocco2014standardized} & 0.77(0.09)           & 0.53 (0.24)          & 0.78 (0.12)          & 0.63 (0.31) \\
		\midrule
		Shadow          & Proposed & \textbf{0.87(0.06)} & \textbf{0.27 (0.25)} & 0.76 (0.12)          & 0.80 (0.45)          \\
		& Faraji et al. \cite{faraji2018segmentation}   & 0.86 (0.07)          & 0.29 (0.20)          & 0.58 (0.13)          & 1.24 (0.39)          \\
		& Downe et al. \cite{downe2008segmentation}    & 0.76 (0.11)          & 0.55 (0.26)          & 0.74 (0.16)          & 0.77 (0.48)          \\
		& Exarchos et al. \cite{balocco2014standardized} & 0.80 (0.10)          & 0.46 (0.19)          & \textbf{0.82 (0.11)} & \textbf{0.57 (0.28)} \\ \bottomrule
	\end{tabular}
\end{table*}

\subsection{The Effectiveness of Data Augmentation}
\label{subsec:eff-data-aug}

We validate the effectiveness of data augmentation with a small experiment. In each case, 5 models with identical configurations are trained and we use the ensemble strategy illustrated in \cite{ciresan2012deep} to produce the final prediction. The result is shown in Table~\ref{table:aug-compare}. Note that this result is based on the predictions produced directly by the ensembling without contour extraction. No matter which type of vessel segmentation the model predicts, we can safely conclude that the augmentation helps to improve the segmentation performance.

\subsection{On Evaluating the Refining Branch}
\label{subsec:eval-refining}

Does refining branch really help? We use the exact same configuration to train two groups of 5 models. One group includes the proposed model, another group is the proposed model without the refining branch. The evaluation procedures and metrics are as same as we did for the data augmentation evaluation, the result is shown in Table \ref{table:tri-compare}. There are, indeed, improvements made by the refining branch.

\subsection{Segmentation Results}
\label{subsec:ivus-exp}

In this section, we present and discuss experimental results on the IVUS dataset \cite{balocco2014standardized}. We train 10 models with the configuration mentioned in Section \ref{subsec:train} and ensemble the predictions followed by contour extraction to produce the final prediction mask. 

The quantitative result is shown in Table \ref{table:experiment}. As we can see, IVUS-Net outperforms existing methods by a significant margin. According to the Jaccard Measure, we achieve 4\% and 8\% improvement for the lumen and the media, respectively. If we look at the Hausdorff distance, IVUS-Net obtains 8\% and 20\% improvement for the lumen and the media, respectively. 

\begin{figure*}[t!]
	\centering
	\includegraphics[width=0.9\textwidth]{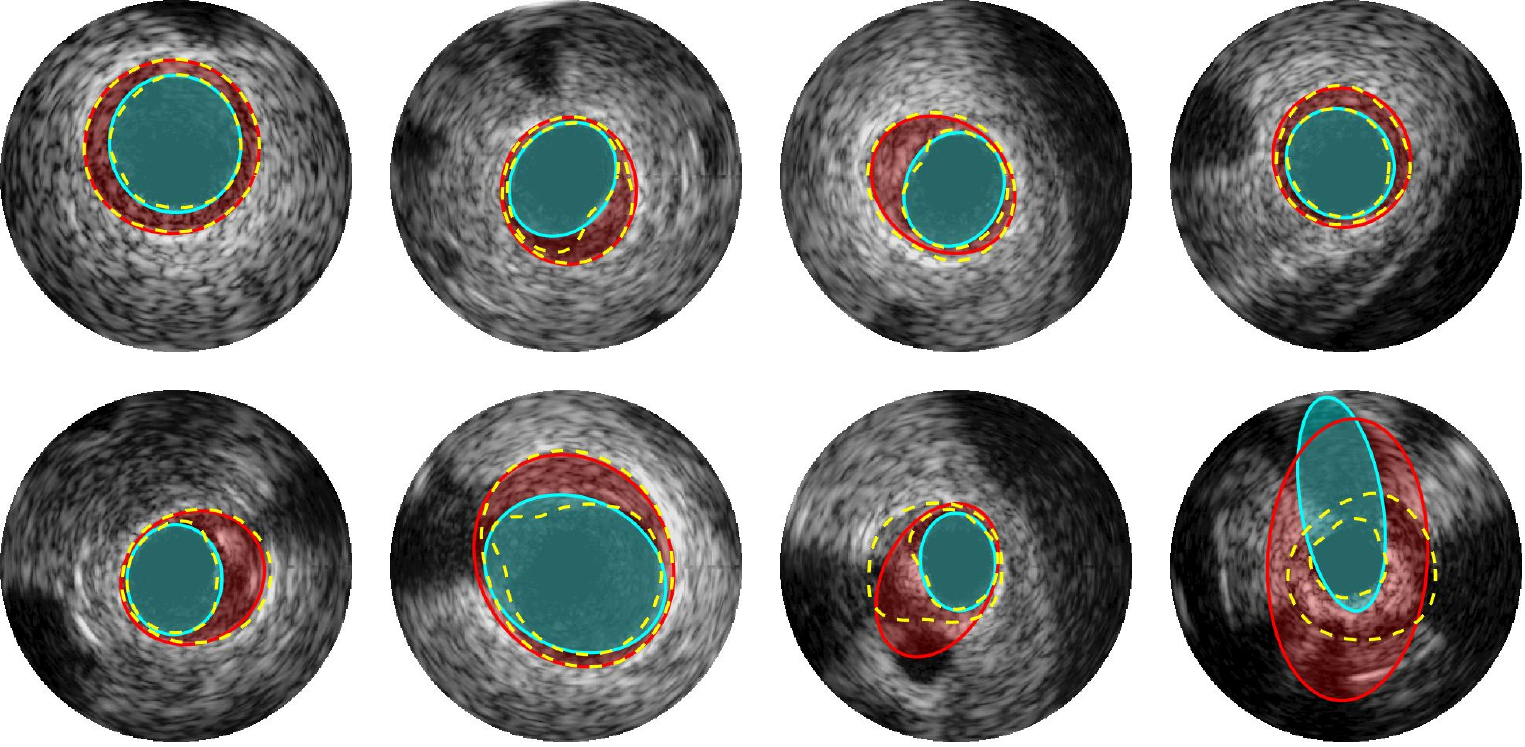}
	\caption{Lumen and media segmentation results. Segmented lumen and media have been highlighted by cyan and red colors, respectively. The yellow dashed lines illustrate the gold standard that have been delineated by four clinical experts~\cite{balocco2014standardized}.}
	\label{fig:testset}
\end{figure*}

IVUS-Net performs particularly well on images with no artifact. Furthermore, it improves the performance by a large margin for segmenting both the lumen and the media according to the Hausdorff distance. The reason why IVUS-Net does not exceed all the methods in every single categories of \cite{balocco2014standardized} can be addressed from two perspectives. First, the training set is too small to capture all the common artifacts in the real world and even the test set. But the architecture is still considerably effective as the training set contains only 1 image with side vessels artifact while the test set contains 93 frames with side vessel artifacts. Secondly, the shadow artifacts are generally overlapped with parts of the media area that makes the segmentation becomes much more challenging since the media regions leak to the background. Some predictions are illustrated in Fig.~\ref{fig:testset}. 


\section{Conclusion}
\label{sec:conclusion}

In this paper, we proposed IVUS-Net for the segmentation of arterial walls in IVUS images as well as a contour extraction post-processing step that specifically fits for the IVUS segmentation task. We showed that IVUS-Net can outperform the existing conventional methods on delineating the lumen and media vessel walls. This is also the first deep architecture-based work that achieves segmentation results that are very close to the gold standard. We evaluated IVUS-Net on a publicly available dataset containing 326 IVUS frames. The results of our evaluation showed the superiority of IVUS-Net output segmentations over the current state-of-the-arts. Also, IVUS-Net can be employed in real-world applications since it only needs 0.15 second to segment any IVUS frame.

\section*{Acknowledgment}
The authors would like to thank the PhD students in the Multimedia Research Centre at University of Alberta. Special thanks to Xinyao Sun for the discussions on the related work and the network architecture figure design. 

{\small
\bibliographystyle{ieee}
\bibliography{refs}
}

\end{document}